\pdfoutput=1

\documentclass[11pt]{article}

\usepackage[]{acl}

\usepackage{times}
\usepackage{latexsym}

\usepackage[T1]{fontenc}

\usepackage[utf8]{inputenc}

\usepackage{microtype}

\usepackage{inconsolata}

\usepackage{graphicx}

\usepackage{amsmath}
\usepackage{amsfonts}
\usepackage{dsfont}
\usepackage{booktabs}
\usepackage{multirow}
\usepackage{tcolorbox}

\newcommand{\factalign}{\textsc{FactAlign}}
\newcommand{\factaligns}{\textsc{FactAlign }}

\title{\factalign: Long-form Factuality Alignment of Large Language Models}

\author{Chao-Wei Huang \quad Yun-Nung Chen \\
        National Taiwan University, Taipei, Taiwan \\
        \texttt{f07922069@csie.ntu.edu.tw} \quad \texttt{y.v.chen@ieee.org}}

\begin{document}
\maketitle
\begin{abstract}
Large language models have demonstrated significant potential as the next-generation information access engines.
However, their reliability is hindered by issues of hallucination and generating non-factual content.
This is particularly problematic in long-form responses, where assessing and ensuring factual accuracy is complex.
In this paper, we address this gap by proposing \factalign, a novel alignment framework designed to enhance the factuality of LLMs' long-form responses while maintaining their helpfulness.
We introduce fKTO, a fine-grained, sentence-level alignment algorithm that extends the Kahneman-Tversky Optimization (KTO) alignment method. 
Leveraging recent advances in automatic factuality evaluation, \factalign~utilizes fine-grained factuality assessments to guide the alignment process.
Our experiments on open-domain prompts and information-seeking questions demonstrate that \factaligns significantly improves the factual accuracy of LLM responses while also improving their helpfulness.
Further analyses identify that \factaligns is capable of training LLMs to provide more information without losing factual precision, thus improving the factual F1 score.\footnote{Our source code, datasets, and trained models are publicly available at \url{https://github.com/MiuLab/FactAlign}.}
\end{abstract}

\section{Introduction}
Generating natural language provides a natural interface for humans to communicate with artificial intelligence.
With the emergence of large language models (LLM)~\cite{brown2020language}, they immediately demonstrate the potential to become the next-generation engine for information access due to their ability to generate long-form natural language response to human queries.
Given the large-scale pre-training on web-scale datasets, LLMs demonstrate impressive capabilities of answering diverse questions, showcasing the vast amount of knowledge they possess.
The post training techniques, i.e., instruction tuning~\cite{wei2022finetuned} and reinforcement learning from human feedback (RLHF)~\cite{ouyang2022training}, further train LLMs to respond in a more human preferable way, e.g., generating coherent and detailed responses.

\begin{figure}[t]
    \centering
    \includegraphics[width=\linewidth]{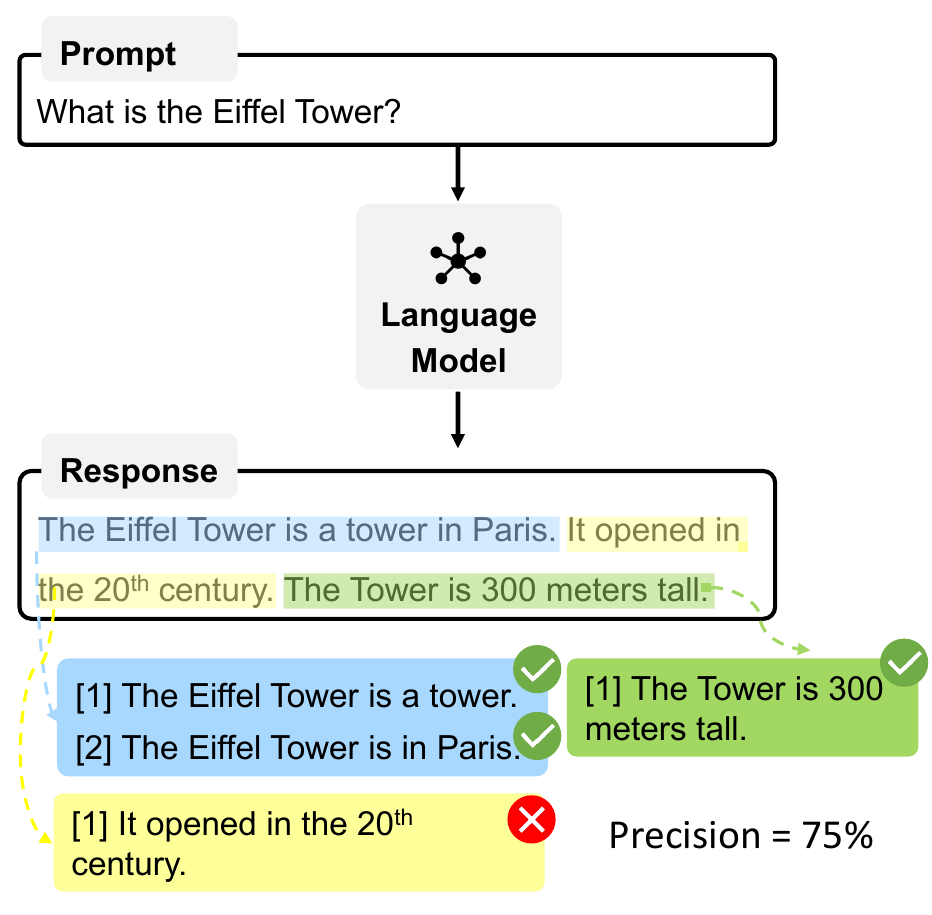}
    \caption{An example of the evaluation long-form factuality. The long-form response is broken down into subclaims and verified separately. The factual precision score can be calculated as the precision of all subclaims.}
    \label{fig:longform}
\end{figure}

Despite their impressive reasoning capabilities and wide-range knowledge, research has shown that LLMs still struggle with hallucination~\cite{xu2024hallucination,rawte-etal-2023-troubling} and generating non-factual content~\cite{min-etal-2023-factscore}.
An example of long-form generation and factuality assessment is illustrated in Figure~\ref{fig:longform}.
These issues hinder the reliability of LLMs and make it hard to be adopted to real-world settings where factual accuracy is a crucial requirement for most applications.
The long-form responses make these issues more complex as it is non-trivial to quantify the level of long-form factuality~\cite{wei2024long}, let alone to improve it.
Meanwhile, most research focuses on improving the helpfulness of LLM chatbots and their reasoning capabilities, with little emphasis on the factuality of the responses.

In this paper, we aim to improve the reliability of LLMs by enhancing the factuality of their long-form responses.
Recent advances of automatic factuality evaluators show that they are capable of providing factuality assessment at the atomic fact level~\cite{min-etal-2023-factscore,wei2024long}.
To leverage those fine-grained factuality assessments, we propose \factalign, an alignment framework designed to improve LLMs' long-form factuality while maintaining the same level of helpfulness.
We introduce a fine-grained alignment algorithm, \textbf{fKTO}, which extends the Kahneman-Tversky Optimization (KTO;~\citet{ethayarajh2024kto}) alignment algorithm to sentence-level.
We conduct experiments on both open-domain prompts and information-seeking questions and demonstrated that our proposed \factaligns can effectively improve long-form factuality of LMs while maintaining their helpfulness.

Our main contributions can be summarized as the following:
\begin{itemize}
    \item We introduce fKTO, a sentence-level alignment algorithm that can leverage fine-grained signals provided by a long-form factuality evaluator.
    \item We propose \factalign, a framework to align LMs with fine-grained signals to generate responses that are more factual, while keeping their helpfulness.
    \item The effectiveness of the proposed components are validated through detailed analyses.
\end{itemize}

\section{Related Work}
\subsection{Language Model Alignment}
Alignment, i.e., aligning language models to human values, has been a very popular research field recently.
Prior work such as InstructGPT~\cite{ouyang2022training} and LLaMA-2~\cite{touvron2023llama} showcased that RLHF~\cite{bai2022training} enhances models' ability to follow instructions significantly.
Fine-grained RLHF~\cite{wu2024fine} proposed to leverage fine-grained rewards for better alignment.
Constituional AI~\cite{bai2022constitutional} and RLAIF~\cite{lee2023rlaif} introduced AI feedback to eliminate the requirement of human annotation.

Another line of research focused on alignment without RL.
DPO~\cite{rafailov2023direct} derived a simple objective for alignment, thus attracting rapid adoption.
KTO~\cite{ethayarajh2024kto} eliminated the requirement of pairwise preference data.
Our proposed alignment algorithm, fKTO, extends KTO to sentence-level, which can leverage the fine-grained signals provided by a long-form factuality evaluator.

\subsection{Factuality of Langage Models}
Factuality and hallucination have been long-standing issues for natural language generation~\cite{lee2022factuality,ji2023survey}.
\citet{lee2022factuality}, \citet{li2023inferencetime}, and \citet{chuang2024dola} proposed decoding techniques that improved factuality of LMs.
\citet{shuster-etal-2021-retrieval-augmentation} reduced hallucination by retrieval-augmented generation.
\cite{dhuliawala2023chain} proposed chain-of-verification to reduce LLM hallucination.
SelfCheckGPT~\cite{manakul-etal-2023-selfcheckgpt} proposed a method to self-check factuality by sampling multiple generations.
FactScore~\cite{min-etal-2023-factscore,chiang2024merging} and LongFact~\cite{wei2024long} both introduced frameworks for evaluating factuality of long-form generations.
FAVA~\cite{mishra2024fine} introduced fine-grained hallucination categories to evaluate the models and provided a detailed view of the hallucination issues of LLMs.
Our proposed method also utilize a long-form factuality evaluator, while focusing on leveraging the provided factuality assessments for better factuality alignment.

Prior work has also worked on training LMs to be more factual.
FactTune~\cite{tian2024finetuning} leveraged FactScore to construct preference pairs and demonstrated improvement on the bio generation task.
FLAME~\cite{lin2024flame} introduced factuality-aware alignment which combines FactTune with open-domain prompts.
KnowTuning~\cite{lyu2024knowtuning} proposed knowledge augmentation which constructs synthetic pairs for DPO training.
On the other hand, recent work has shown that fine-tuning LMs on new knowledge might encourage hallucinations~\cite{gekhman2024does,kang2024unfamiliar}.
Our work additionally proposes fKTO for fine-grained factuality alignment, which achieves superior performance.

\section{Preliminaries}
In this paper, we aim to improve the long-form factuality of LLMs by factuality alignment.
In this section, we introduce an overview of the task of long-form factuality and alignment algorithms.

\subsection{Long-form Factuality}
LLMs excel at generating long-form responses with detailed description and explanation.
However, evaluating the factuality of long-form generations is non-trivial.
In this paper, we define the factuality score of a long-form response as an aggregation of the factuality score of each individual atomic fact, following FactScore~\cite{min-etal-2023-factscore} and LongFact~\cite{wei2024long}.
More formally, given a knowledge corpus $\mathcal{C}$, an user prompt $x$ and the response $y = \mathcal{M}(x)$ generated by a model $\mathcal{M}$, we first decompose $y$ into atomic statements $A = \{a_1, \cdots, a_{|A|}\}$.
For each atomic statement $a_i$, its factuality score $f(a_i)$ is defined as whether it is supported by the knowledge in $\mathcal{C}$, i.e., $f(a_i) = \mathds{1}[a_i\text{ is supported by } \mathcal{C}]$.
Then, the factuality score of the long-form response $y$ can be defined as $f_{\mathcal{A}}(y) = \mathcal{A}(\{f(a_1), \cdots, f(a_{|A|})\})$, where $\mathcal{A}$ is an aggregation function that can be defined in various ways.

In this paper, we adopt two metrics for long-form factuality: factual precision as defined in FactScore~\cite{min-etal-2023-factscore} and factual f1 score as defined in LongFact~\cite{wei2024long}.
Factual precision measures the overall precision of the atomic statements:
\begin{equation*}
    f_{prec}(y) = \frac{\sum_{i=1}^{|S|}{f(a_i)}}{|A|}.
\end{equation*}
While factual precision is simple, it could be easily exploited.
A model could obtain a very high factual precision score by only generating one statement that has the highest confidence.

On the other hand, factual f1 assumes that a certain amount of information is desired by the user and additionally considers the factual recall:
\begin{equation*}
    f_{f1\text{@}K}(y) =
        \begin{cases}
          \frac{2 \cdot f_{prec}(y) \cdot f_{rec\text{@}K}(y)}{f_{prec}(y) + f_{rec\text{@}K}(y)} & \text{if } |A| > 0 \\
          0 & \text{if } |A| = 0,
        \end{cases}
\end{equation*}
where $f_{rec\text{@}K}(y) = \text{min}(1.0, \frac{|A|}{K})$ is the factual recall score assuming that at least $K$ statements are desired by the user.
Factual f1 is less exploitable than factual precision as it punishes the model when it only generates few statements.

\subsection{Kahneman-Tversky Optimization}
Training LLMs that are aligned to human values typically involves three stages: 1) pre-training, 2) supervised fine-tuning, and 3) reinforcement learning from human feedback (RLHF).
The first two stages maximize the sequence generation likelihood of the LM given a dataset of either diverse pre-training data or human-annotated instruction-following data.
The third stage, RLHF, aims to maximize the expected reward of LM generations, where the reward usually is defined as human preferences~\cite{ouyang2022training}.
As a result, the RLHF stage enables LMs to generate responses that are more preferable by humans, which is vital for creating intelligent assistants.

While the success of the RLHF framework is eminent, its adoption is hindered by the complexity of the framework, the unstability of the training process, and the increased training time due to the requirement of online sample generation.
To this end, prior work has proposed alignment algorithms that do not require RL, thus attracting mass adoption.
Direct Preference Optimization (DPO;~\citet{rafailov2023direct}) derives a simpler objective from the RLHF, eliminating the requirement of a reward model and the RL optimization process.
More recently, \citet{ethayarajh2024kto} introduced Kahneman-Tversky Optimization (KTO), which derives a family of human-aware alignment loss functions.
The objective function of KTO is even simpler than DPO.
It only requires a binary label for each prompt-response pair $(x, y)$, as opposed to DPO which requires pairwise preference labels for each triplet $(x, y_1, y_2)$.
This relaxed data requirement enables us to extend the algorithm to sentence-level, which we will discuss in Section~\ref{sec:alignment}.
More formally, the KTO loss is defined as:
\begin{equation*}
    \mathcal{L}_{\text{KTO}} = \frac{1}{|\mathcal{B}|} \sum_{x,y \in \mathcal{B}} (\lambda_y - v(x, y)),
\end{equation*}
where $\mathcal{B}$ denotes the minibatch, $\lambda_y$ denotes the weight of the chosen and rejected samples, and
\begin{align*}
    v(x, y) &=
        \begin{cases}
            \lambda_c \sigma(\beta (r_{\theta}(x, y) - z_0)) \text{ if } c(x, y) = 1, \\
            \lambda_r \sigma(\beta (z_0 - r_{\theta}(x, y))) \text{ if } c(x, y) = 0,
        \end{cases} \\
    z_0 &= \mathbb{E}_{y' \sim \mathcal{D}} [\text{KL} ({\pi_{\theta}(y' \mid x')} \Vert {\pi_{\text{ref}}(y' \mid x')})], \\
    r_{\theta}(x, y) &= \log \frac{\pi_{\theta}(x, y)}{\pi_{\text{ref}}(x, y)},
\end{align*}
where $c(x, y)$ denotes the preference function, i.e., $c(x, y) = 1$ if the response $y$ is \textit{chosen}.
\citet{ethayarajh2024kto} demonstrated that KTO achieves on par or better alignment performance compared to DPO.
KTO also works well under the scenario where the number of chosen and rejected samples are significantly unbalanced, e.g., 1:9.

\begin{figure*}[ht]
    \centering
    \includegraphics[width=\textwidth]{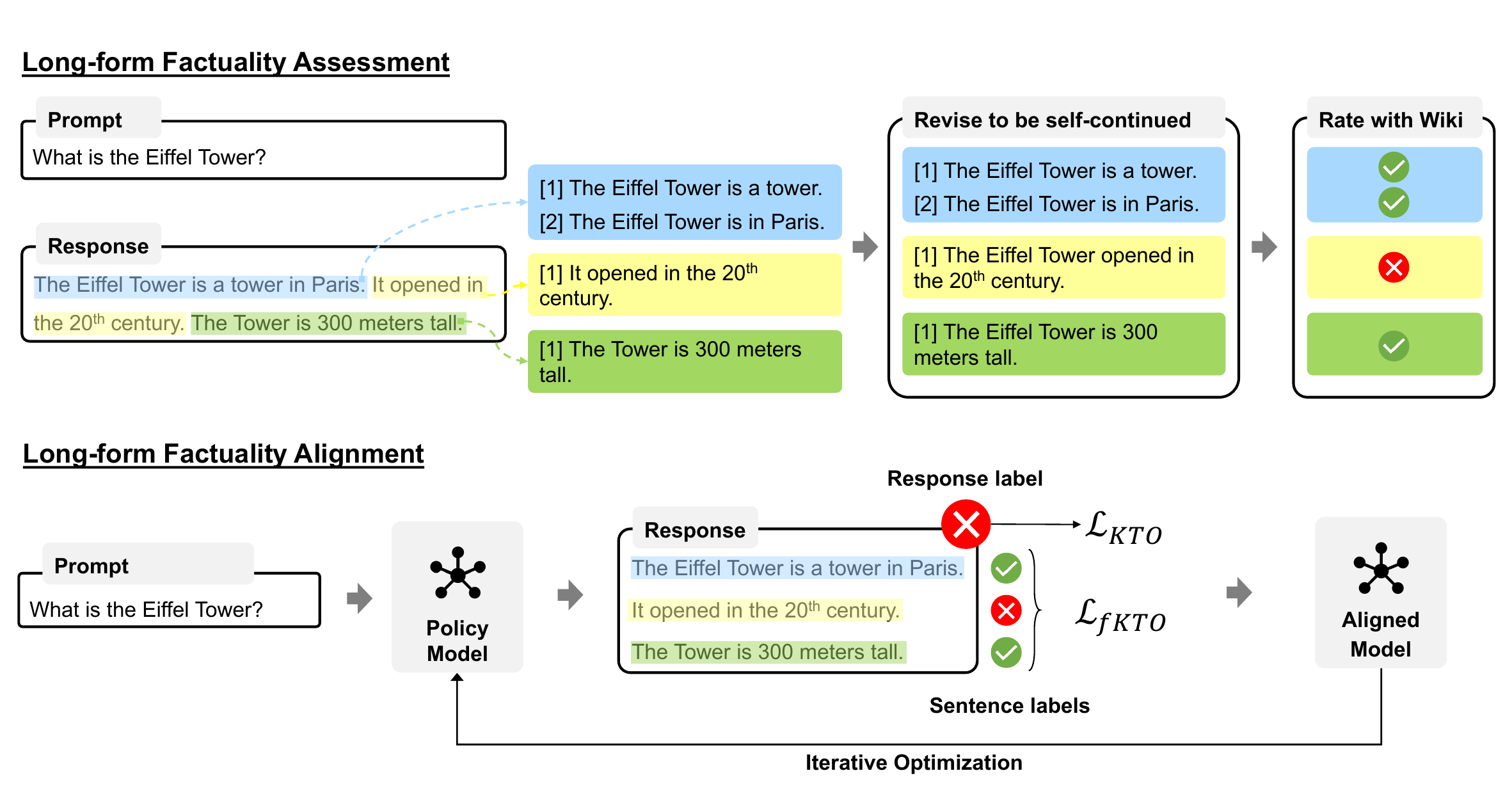}
    \caption{An overview of our \factaligns framework. Top: the pipeline for long-form facutality assessment. Bottom: the long-form factuality alignment process.}
    \label{fig:overview}
\end{figure*}

\section{\factalign: Aligning Language Models for Long-form Factuality}
In this section, we introduce our proposed framework \factalign.
An overview of our framework is illustrated in Figure~\ref{fig:overview}.

\subsection{Automatic Long-form Factuality Evaluator}
Obtaining fine-grained factuality annotations for long-form responses by human annotation is very costly.
For example, \citet{min-etal-2023-factscore} estimated that evaluating one generation costs \$4.
In this work, we employ an automatic factuality evaluator for long-form responses.
The factuality evaluator, following the design of FactScore~\cite{min-etal-2023-factscore} and SAFE~\cite{wei2024long}, is a workflow of 4 stages: 1) atomic statement decomposition, 2) query generation, 3) relevant knowledge search, and 4) final factuality assessment.
Note that stage 2 and 3 can be run multiple times to enrich the searched knowledge.

\paragraph{Atomic Statement Decomposition}
The response $y$ is first split into sentences $S = \{s_1, \cdots, s_{|S|}\}$, and each sentence is decomposed into atomic facts $A$.
We add an additional step to revise the decomposed atomic statements into self-contained statements $s_i'$ with \textsc{GPT-3.5-Turbo} following SAFE.

\paragraph{Query Generation}
We prompt \textsc{GPT-3.5-Turbo} to generate a search query given the revised statement $s_i^{'}$ and possibly the previously generated queries and found knowledge snippets.

\paragraph{Relevant Knowledge Search}
We employ Wikipedia as the knowledge corpus $\mathcal{C}$ following FactScore.
While the coverage of Wikipedia is more limited compared to commercial search engines like Google Search, we opt for Wikipedia as this reduces cost and allows us to fully manage the knowledge search component under a controlled setting.
We perform search with the generated query and obtain the top-k most relevant knowledge snippets.

\paragraph{Final Factuality Assessment}
We prompt \textsc{GPT-3.5-Turbo} to provide the final factuality assessment of a revised statement $s_i^{'}$, which is either \textit{Supported} if the statement is supported by the knowledge snippets, or \textit{Not Supported} otherwise.
The statement-level score is then defined as $f(a_i) = \mathds{1}[a_i \text{ is Supported}]$.
Note that $f(a_i)$ represents whether the statement is supported with respect to Wikipedia, not whether it is globally true.

\subsection{Long-form Factuality Alignment}
\label{sec:alignment}
At the core of the \factaligns framework is the alignment algorithm, which operates on two granularities: response-level and sentence-level.

\subsubsection{Response-level Alignment}
We employ the standard KTO loss $\mathcal{L}_{\text{KTO}}$ for response-level alignment.
The preference labels $c(x, y)$ in the KTO loss can be defined and obtained in various ways.
For instance, most prior work utilized human-annotated preference labels or pseudo labels provided by LLMs.
In order to align for factuality, we treat a response $y$ as a \textit{chosen} sample if the factual f1 score of the response is greater than a threshold $t$:
\begin{equation*}
    c(x, y) = \mathds{1}[f_{f1\text{@}K}(y) > t].
\end{equation*}
By minimizing the response-level loss, we align the LMs to generate responses that have higher factual f1 scores.

In addition to the data for factuality alignment, the response-level loss is compatible to other forms of preference data.
For example, in order to make the model more helpful, we can include diverse preference datasets that are based on human preferences.
In practice, we include general-domain alignment datasets during training to make sure the model is aligned to diverse human values.

\subsubsection{Sentence-level Alignment}
Since our factuality evaluator provides assessments at a finer granularity, we propose a fine-grained alignment algorithm, \textbf{fKTO}, to leverage these signals by extending the KTO alignment algorithm to sentence-level.
The fKTO loss is defined as
\begin{equation*}
    \mathcal{L}_{\text{fKTO}} = \frac{1}{|\mathcal{B}|} \sum_{x,y \in \mathcal{B}} \frac{1}{|S|} \sum_{i=1}^{|S|} (\lambda_f - v(x \mathbin\Vert s_{<i}, s_i)),
\end{equation*}
where $x \mathbin\Vert s_{<i}$ denotes the concatenation of $x$ and $s_{<i}$  which denotes sentences before $s_i$.
In this objective function, a sentence $s_i$ is treated as the completion given $x \mathbin\Vert s_{<i}$.
A sentence is \textit{chosen} if the average precision of its atomic statements is higher than a threshold $t_s$.
\begin{equation*}
    c(x \mathbin\Vert s_{<i}, s_i) = \mathds{1}\Bigg[ \frac{\sum_{j=1}^{|A_{s_i}|}f(a_j)}{|A_{s_i}|} > t_s \Bigg],
\end{equation*}
where $A_{s_i} = \{a_j \mid a_j \in s_i\}$ denotes the atomic statements in sentence $s_i$.
The sentence-level loss provides training signals at a finer-grained level, thus enabling the model to be aligned more effectively.
Note that the relaxed data requirement enables KTO to be easily extended to the sentence-level, as opposed to algorithms that require pairwise preference labels, e.g., DPO.

Finally, the loss function we optimize is the combination of the response-level and sentence-level losses:
\begin{equation*}
    \mathcal{L} = \mathcal{L}_{\text{KTO}} + \lambda \cdot \mathcal{L}_{\text{fKTO}},
\end{equation*}
where $\lambda$ is the weight of the sentence-level loss.

\subsection{Iterative Optimization}
\label{sec:iter}
With the alignment algorithms introduced above, we can align LMs to be more factual and more helpful.
However, the responses and factuality assessments are obtained in an offline fashion, i.e., we sample the responses and their factuality labels before training the model and use this data throughout training.
This creates a discrepancy between the assessed responses and the model being trained, which would hinder the alignment process due to distributional shift.
Hence, we employ an iterative optimization procedure, where we periodically sample new responses with the trained model and assess their factuality.
The newly generated responses are then included in the training dataset for the next iteration.

\section{Experimental Stetup}
We conduct experiments to validate the effectiveness of our proposed framework \factalign.
Furthermore, we perform analyses to discuss the effectiveness of each component in the framework.

\subsection{Datasets}
\paragraph{Supervised Fine-tuning (SFT)}
We employ the Deita dataset~\cite{liu2024what} for supervised fine-tuning before performing alignment to ensure basic instruction-following capabilities of the model.
The Deita dataset consists of high-quality data selected from UltraChat~\cite{ding2023enhancing}, ShareGPT\footnote{\url{https://sharegpt.com}}, and WizardLM~\cite{xu2024wizardlm}.

\paragraph{General-domain Alignment}
We follow the Zephyr recipe~\cite{tunstall2023zephyr} and employ the UltraFeedback dataset~\cite{cui2023ultrafeedback} as the general-domain alignment dataset.
The UltraFeedback consists of prompts across multiple domains and completions generated from multiple LLMs to enrich diversity.
We use the binarized version of the dataset\footnote{\url{https://huggingface.co/datasets/HuggingFaceH4/ultrafeedback_binarized}} and decouple the pairs for the KTO loss.

\paragraph{Factuality Alignment}
We generate information-seeking prompts following the data creation procedure from LongFact~\cite{wei2024long}.
LongFact consists of 38 topics chosen to ensure diverse coverage.
For each topic, we generate 30 prompts with \textsc{GPT-4-Turbo} and sample generations with our policy model.
The generations are then assessed by the long-form factuality evaluator and labeled with factuality assessments at an atomic statement level.
For each iteration of iterative optimization, we generate a new set of prompts and sample generations with the currently aligned model.

\subsection{Long-form Factuality Evaluator}
We employ \texttt{gpt-3.5-turbo} to perform atomic statement decomposition, query generation, and final factuality assessment.
The generation temperature is set to 0.1.
We use the preprocessed Wikipedia corpus from the Dec. 20, 2021 dump released by \citet{10.5555/3648699.3648950} as our knowledge corpus $\mathcal{C}$, which consists of 33 million passages.
A pre-trained retriever ColBERT-v2~\cite{santhanam-etal-2022-colbertv2} is used to encode all passages and perform retrieval given a query.
We retrieve top-3 passages for each query and combine them with the previously retrieved passages for final factuality assessment.
At most 2 search steps are performed to retrieve relevant passages for each statement.
Detailed prompts can be found in Appendix~\ref{app:prompts}.

\subsection{Models}
We employ the pre-trained \texttt{gemma-2b} model~\cite{team2024gemma} as our policy model, which is an open-weight model pre-trained on large-scale datasets across diverse domains.
The model is first finetuned with the Deita SFT dataset, and then aligned with the alignment datasets.

We also conduct experiments on \texttt{LLaMA-3 8B}~\cite{meta2024introducing} and \texttt{Phi3-Mini} models~\cite{abdin2024phi}, which are both open-weight models which were aligned with proprietary data.

\subsection{Evaluation Procedure}
The trained models are evaluated on two aspects: \textit{long-form factuality} and \textit{helpfulness}.

\paragraph{Long-form Factuality Evaluation}
We evaluate models' long-form factuality following the procedure of SAFE~\cite{wei2024long}\footnote{\url{https://github.com/google-deepmind/long-form-factuality}}.
We choose the \textit{LongFact-object} subset following the original work, which consists of 38 topics.
We change the Google Search API to our Wikipedia retriever due to resource and budget constraint.
In preliminary experiments, we find that this change have very little impact on the evaluation outcome.
Our evaluator has correlation scores of 0.93 and 0.82 with SAFE for the number of \textit{Supported} and \textit{Not Supported} assessments, respectively.
We follow SAFE to add an postamble to each prompt to ask for the model to generate as many details and examples as possible.
We report $f1\text{@}100$ as the main evaluation metric.
We also report the factual precision and factual recall scores.
In addition, we evaluate models with FactScore~\cite{min-etal-2023-factscore}.
We run the evaluation from its official implementation\footnote{\url{https://github.com/shmsw25/FActScore}} and use \textsc{GPT-3.5-Turbo} as the evaluator instead of InstructGPT.
FactScore can be interpreted as the factual precision of bio generation.

\paragraph{Helpfulness Evaluation}
We evaluate models' helpfulness on MT-Bench~\cite{zheng2023judging}, a popular benchmark that includes challenging multi-turn open-ended questions for evaluating chat assistants.
The automatic judgement is performed by \textsc{GPT-4} with a score of 1 to 10, which is shown to be highly-correlated with human judgement.
The evaluation is done with their official implementation\footnote{\url{https://github.com/lm-sys/FastChat/tree/main/fastchat/llm_judge}}.

\subsection{Implementation Details}
We set the threshold $t$ to 0.75, meaning that the response is chosen if its f1@100 is higher than 0.75.
The threshold for sentences $t_s$ is set to 1.0, i.e., the sentence is only chosen if all of its atomic statements are supported.
During training, we set $\beta = 0.1$ for KTO and $\beta_f = 0.5$ for fKTO.
The weight of $\mathcal{L}_{\text{fKTO}}$, $\lambda$, is set to 2.0.
The learning rate is set to 5e-7 with a linear learning rate schedule.
We set the effective batch size to 16 and train for 1 epoch for each iteration.
In order to reduce GPU memory consumption during training, we optimize the model with the 8-bit version of the AdamW optimizer.
We iteratively optimize the LM as described in Section~\ref{sec:iter} for 3 iterations.
All experiments are run on 4xV100 GPUs.
Each training run takes 1 to 2 hours to finish.
We estimate that each evaluation run costs \$25 in API credits.

\begin{table*}[t!]
\centering
\begin{tabular}{l|ccc|cc|c}
\toprule
\textbf{}     & \multicolumn{3}{c|}{\textbf{LongFact}}                                                         & \multicolumn{2}{|c|}{\bf FactScore}                               & \multicolumn{1}{|c}{\multirow{2}{*}{\bf MT-Bench}} \\
              & \multicolumn{1}{c}{$f1\text{@}100$} & \multicolumn{1}{c}{Precision} & \multicolumn{1}{c|}{\# claims} & \multicolumn{1}{c}{FS}    & \multicolumn{1}{c|}{\# claims} & \multicolumn{1}{c}{}                          \\
\midrule
GPT-4-Turbo   & 91.85                       & 87.99                         & 108.9                           & 83.98 & 49.7      & 8.99                                          \\
GPT-3.5-Turbo & 61.59                       & 89.25                         & 52.4                            & 75.79 & 18.1        & 7.94                                          \\
Olmo-7B-Instruct & 75.42 & 66.8 & 131.6 & - & - & - \\
LLaMA-2-70B-Chat   & 83.78 &	79.33 &	112.5 & 59.06 & 65.4 & 6.85   \\
\midrule
Gemma-2B-SFT  & 61.74 & 77.41 & 66.8 & 57.02 & 27.4 & 4.41 \\
\multicolumn{1}{r|}{+ \factalign}    & \bf 86.51 & \bf 79.59 & \bf 135.1 & \bf 59.77 & \bf 51.7 & \bf 5.70       \\
\midrule
Phi3-Mini  & 66.91 & 74.41 & 81.5 & 58.33 & 36.7 & 6.58 \\
\multicolumn{1}{r|}{+ \factalign}    & \bf 82.31 & \bf 78.29 & \bf 106.7 & \bf 61.21 & \bf 49.6 & \bf 6.63       \\
\midrule
LLaMA3-8B-Chat  & 79.98 & 71.89 & 121.5 & 54.96 & 50.6 & 7.38 \\
\multicolumn{1}{r|}{+ \factalign}    & \bf 87.32 & \bf 78.46 & \bf 132.6 & \bf 62.84 & \bf 55.7 & \bf 7.44      \\
\bottomrule
\end{tabular}
\caption{Main results of our experiments. FS denotes the FactScore and \# claims denotes the average number of claims. We report percentage points for $f1\text{@}100$, precision, and FS. We mark the best scores among the Gemma-2B models in bold.}
\label{tab:main}
\end{table*}

\section{Results}
We present the main results in Table~\ref{tab:main}, 
where we contrast \factaligns with both proprietary models (GPT-4-Turbo and GPT-3.5-Turbo), a prominent open-weight model (LLaMA-2-70B-Chat~\cite{touvron2023llama}), and a fully open-source model (Olmo-7B-Instruct)~\cite{groeneveld2024olmo}.
The comparison involves our baseline model, the Gemma-2B model\footnote{\url{https://huggingface.co/google/gemma-2b}}, which has been fine-tuned using our SFT dataset, Deita. This model serves as the foundational policy model for all subsequent aligned models. Additionally, we benchmark against the rejection sampling fine-tuning method~\cite{yuan2023scaling}, involving supervised fine-tuning with selected samples from our alignment dataset. This method shows modest improvements.

Remarkably, our \factaligns framework significantly improves the long-form factuality and helpfulness of the baseline model, achieving relative improvements of 40.1\% and 29.2\% in terms of $f1\text{@}100$ and average score on MT-Bench, respectively. 
These results demonstrate our capability to simultaneously refine LMs for enhanced factuality and utility. Moreover, \factaligns also boosts the FactScore of the baseline models and outperforms larger models like GPT-3.5-Turbo and LLaMA-2-70B-Chat in both $f1\text{@}100$ and FactScore metrics. 
This demonstrates the potential for smaller LMs, through precise alignment, to surpass general-domain large LMs in factual accuracy.

With a detailed examination of the metrics, it is evident that \factaligns primarily improves factual recall, increasing the output of factual claims from 66.8 to 135.1, while slightly improving factual precision from 77.41 to 79.59. This enhancement suggests that \factaligns primarily amplifies output volume while maintaining factual precision. This trend echoes findings from general-domain alignment research, which indicates that alignment algorithms typically promote longer outputs, likely due to a combined human and LM preference for more extensive responses~\cite{dubois2024length}. A qualitative example of this can be found in Appendix~\ref{app:example}.

\begin{table}[t]
\centering
\begin{tabular}{l|c|c}
\toprule
& \textbf{f1@100} & \textbf{Precision}  \\
\midrule
\factalign    & \bf 86.51 & \bf 79.59 \\
\quad - Iterative Optimization & 77.10 & 78.44 \\
\qquad - fKTO    & 73.12   & 73.27 \\
\qquad - General Dataset & 61.33 & 65.72  \\
\qquad - Factuality Dataset & 68.86 & 69.93  \\
Rejection Fine-tuning & 68.33 & 77.86 \\
\bottomrule
\end{tabular}
\caption{Ablation study on LongFact (\%).}
\label{tab:ablation}
\end{table}

\subsection{Ablation Study}
To validate the effectiveness of our proposed components, we conduct an ablation study to understand their contribution to the final improvement.
The results are reported under \factaligns in Table~\ref{tab:ablation}.

Firstly, we remove the iterative optimization technique, where we only perform 1 iteration of training.
As shown in the results, removing iterative optimization significantly degrades the performance, where $f1\text{@}100$ drops by over 10 points.
This result demonstrates that it is crucial to perform iterative optimization or online sampling in order to achieve better performance.
We also observe that training on the same dataset for multiple epochs yields worse performance, showcasing that the alignment data quickly becomes stale and no longer is a good sample after 1 epoch of training.
Note that for all other ablation experiments, we also only performs 1 iteration of training.

\begin{table}[t]
\centering
\begin{tabular}{l|c|c}
\toprule
& \textbf{Seen} & \textbf{Unseen}  \\
\midrule
Gemma-2B-SFT      & 61.97 & 61.36  \\
General-domain only & 69.45 & 68.23 \\
Seen topics only & 76.49 & 72.23 \\
\bottomrule
\end{tabular}
\caption{Performance on seen and unseen topics (\%). We report the f1@100 score on LongFact.}
\label{tab:generalization}
\end{table}

\begin{table}[t]
\centering
\begin{tabular}{c|c|c}
\toprule
$\beta_f$ & threshold $t$ & $f1\text{@}100$  \\
\midrule
0.5 & 0.65 & 74.32 \\
0.5 & 0.75 & 77.10 \\
0.5 & 0.85 & 73.66 \\
0.1 & 0.75 & 75.86 \\
1.0 & 0.75 & 75.12 \\
\bottomrule
\end{tabular}
\caption{Performance with various number of $\beta_f$ and threshold $t$ (\%).}
\label{tab:hyperparam}
\end{table}

Next, we remove the fKTO loss $\mathcal{L}_{\text{fKTO}}$ and align the model with only $\mathcal{L}_{\text{KTO}}$.
Without $\mathcal{L}_{\text{fKTO}}$, the factual f1 score degrades by 4 points from 77.10 to 73.12, demonstrating that the proposed fine-grained alignment objective fKTO can align LMs more effectively.
Note that we observe that the fKTO loss occasionally makes the training process unstable.
We hypothesize that this is due to the amount of factuality data being much less than the general-domain data, thus making the instances with fine-grained label sparse during training.
Hence, the estimation of the fKTO loss becomes slightly unstable.
We will also discuss the sensitivity to hyperparameters in Section~\ref{sec:hyperparams}.

We also conduct an experiment where we exclude the general-domain alignment dataset from our training data.
The performance degrades significantly on all datasets after removing the general-domain alignment dataset.
Upon further investigation, we observe that without general-domain data, LMs easily overfit and often generate repetitive outputs.
This result indicates that a mixture of general-domain datasets and factualy-specific datasets is important to maintain a balance and prevent catastrophic forgetting.

Finally, we exclude the factuality dataset during training, i.e., only align the LM on general-domain datasets.
As shown in the results, aligning with general-domain dataset also improves the long-form factuality and helpfulness of the baseline model.
This indicates that factuality might be encoded in the diverse array of human values present in the general-domain alignment dataset.
However, including the factuality dataset sill achieves significantly superior performance for long-form factuality.

\subsection{Generalization to New Topics}
Since the training data is created with the same set of topics in LongFact, all the topics should be considered \textit{seen} during evaluation.
Note that prompts used in evaluation are excluded during training.
To validate whether \factaligns could generalize to unseen topics, we conduct an additional experiment where we split the topics into 19 seen topics and 19 unseen topics.
We only include the data from the seen topics during training and perform evaluation on the unseen topics.
The results are reported in Table~\ref{tab:generalization}.
The results show that \factaligns performs slightly worse on unseen topics.
Nonetheless, it still outperforms the baseline models significantly, showcasing that the alignment can generalize to unseen topics.

\begin{figure}[t]
    \centering
    \includegraphics[width=\linewidth]{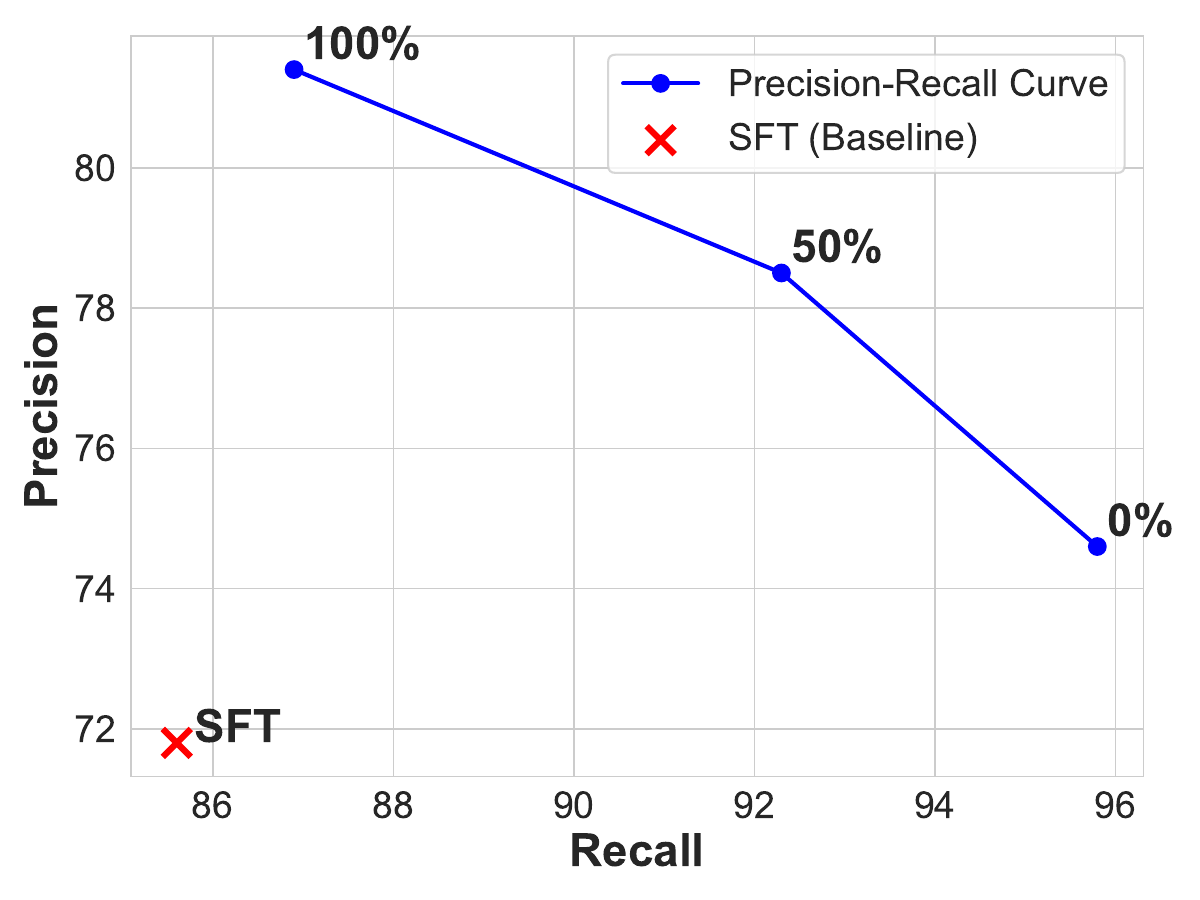}
    \caption{The precision-recall curve with varying ratios of data mixture. SFT denotes the supervised fine-tuned baseline. The labels denote the ratio of the precision data points used.}
    \label{fig:prcurve}
\end{figure}

\subsection{Relationship of Precision-Recall}
By varying the ratio of data points using precision as the threshold and those using recall, we can control the tradeoff between the precision score and the recall score.
We train models with different data mixture and plot the corresponding precision-recall curve in Figure~\ref{fig:prcurve}.
The model trained with 100\% precision data achieves the highest precision score, and the model trained with 100\% recall data achieves the highest recall score.
Furthermore, we can achieve a specific level of factual precision and recall scores on the curve by changing the ratio.
This result demonstrates that \factaligns enables control over the desired factual precision and recall scores.

\subsection{Sensitivity of Hyperparameters}
\label{sec:hyperparams}
We report the performance of \factaligns under various hyperparamter settings.
The results are reported in Table~\ref{tab:hyperparam}.
We observe that the threshold $t$ affects performance slightly, with 0.75 being the best setting.
We also notice that with $t = 0.75$, the labels are balanced, i.e., the number of chosen samples is roughly equal to the number of rejected samples.
This indicates that constructing a balanced dataset perform better for our alignment algorithm.

We also vary the hyperparameter $\beta_f$ and notice that it degrades performance slightly.
Note that the best $\beta_f$ value is higher than the $\beta$ value typically set for KTO, i.e., 0.1.
Our hypothesis is that since fKTO operates on the sentence-level, the log probability difference naturally has a lower magnitude compared to the response-level case.
Thus, a higher value of $\beta_f$ is needed to promote the fine-grained loss to a similar level as the response-level loss.

\section{Conclusion}
In this paper, we address the issue of long-form actuality in LLMs by proposing a novel alignment framework, \factalign.
Our approach, which incorporates a proposed data construction process alongside the fine-grained alignment algorithm fKTO, significantly enhances the factuality of LLMs over long-form responses, while also boosting their helpfulness. 
Our analysis demonstrates that \factaligns enables detailed control over the desired level of factual precision and recall scores.
We believe that the insights and methodologies presented in our work can motivate further advancements in the factuality alignment of LLMs.

\section*{Limitations}
Our work focuses on the factuality aspect of LLMs, which we define as whether the generated response is supported by retrieved evidence.
This definition makes the performance dependent to the performance of the retriever and the coverage of the knowledge corpus.
Moreover, our data creation and evaluation pipeline rely on automatic factuality evaluators.
Even though prior work has validated the effectiveness of these evaluators by showing high correlation with human judgements, the automatic evaluators inevitably might make incorrect judgements.

While \factaligns significantly improves the factuality of LLMs, they still are prone to generate non-factual content.
A calibration method would be complimentary to our method to ensure the reliability of LLMs.

We focus on a controlled setting where the information-seeking prompts are all questions about a certain object.
This is to ensure the reliability of the automatic evaluation process.
Future work could extend the coverage of the information-seeking prompts to more diverse user queries.

\section*{Acknowledgements}
We thank the reviewers for their insightful comments.
This work was financially supported by the National Science and Technology Council (NSTC) in Taiwan, under Grants 111-2222-E-002-013-MY3 and 112-2223-E002-012-MY5. 
We thank to National Center for High-performance Computing (NCHC) of National Applied Research Laboratories (NARLabs) in Taiwan and Google PaliGemma Academic Program for providing computational and storage resources.

\bibliography{custom}

\appendix

\section{Prompts Used}
\label{app:prompts}

We use the following prompt for new prompt generation

\begin{tcolorbox}[width=\columnwidth,colback=white]
\small
\begin{verbatim}
Instructions:
1. Ask a general question about a specific
object (such as a person, place, event, act,
company etc.).
2. The object should be very specific and
niche within the topic of {topic}.
4. Do not ask about any particular aspect
of the object (such as history,
significance, impact, role, purpose, etc.).
5. Follow the question styles in the
provided examples.
6. Wrap the question in square brackets.
\end{verbatim}
\end{tcolorbox}

The following prompt is used for query generation:

\begin{tcolorbox}[width=\columnwidth,colback=white]
\small
\begin{verbatim}
Instructions:
1. You have been given a STATEMENT and
some KNOWLEDGE points.
2. Your goal is to try to find evidence
that either supports or does not
support the factual accuracy of the
given STATEMENT.
3. To do this, you are allowed to
issue ONE search query that you think 
will allow you to find additional
useful evidence.
4. Your query should aim to obtain
new information that does not appear 
in the KNOWLEDGE. This new information
should be useful for determining the
factual accuracy of the given STATEMENT.
5. Format your final query by putting
it in a markdown code block.

KNOWLEDGE:
{_KNOWLEDGE_PLACEHOLDER}

STATEMENT:
{_STATEMENT_PLACEHOLDER}
\end{verbatim}
\end{tcolorbox}

The following prompt is used for final answer assessment:

\begin{tcolorbox}[width=\columnwidth,colback=white]
\small
\begin{verbatim}
Instructions:
1. You have been given a STATEMENT and
some KNOWLEDGE points.
2. Determine whether the given STATEMENT
is supported by the given KNOWLEDGE.
The STATEMENT does not need to be explicitly
supported by the KNOWLEDGE, but should be
strongly implied by the KNOWLEDGE.
3. Before showing your answer, think
step-by-step and show your specific reasoning.
As part of your reasoning, summarize the main
points of the KNOWLEDGE related to the
STATEMENT.
4. If the STATEMENT is supported by the
KNOWLEDGE, be sure to show the supporting
evidence.
5. After stating your reasoning, restate
the STATEMENT and then determine your final
answer based on your reasoning and the
STATEMENT.
6. Your final answer should be either
"Supported" or "Not Supported". Wrap your
final answer in square brackets.
\end{verbatim}
\end{tcolorbox}

\begin{table*}[t!]
\centering
\small
\begin{tabular}{lp{10cm}cc}
\toprule
\multicolumn{1}{c}{\textbf{Model}} & \multicolumn{1}{c}{\bf Response} & \multicolumn{1}{c}{\bf \# Claims} & \multicolumn{1}{c}{\bf $f1\text{@}100$} \\
\midrule
\multicolumn{4}{l}{\textbf{Prompt}: Could you provide some details about the Hague Conventions?} \\
\midrule
Gemma-2B-SFT & Sure! The Hague Conventions are a set of international treaties that were designed to protect the rights of civilians in international armed conflicts. Here are some specific details and examples related to the Hague Conventions:

1. The First Hague Convention: The First Hague Convention was signed in 1899 and was designed to protect the rights of civilians in international armed conflicts. It was signed by 31 countries and established the principle that civilians should not be targeted by military forces.

Example: In 1904, during the Boxer Rebellion in China, British and French troops were sent to suppress the rebellion. However, they were ordered to avoid killing or injuring the Chinese civilians. The British and French troops complied with this principle and did not fire on the civilians.

2. The Second Hague Convention: The Second Hague Convention was signed in 1907 and established the principle that the use of prohibited weapons in war is prohibited. It was signed by 33 countries and established the principle that the use of prohibited weapons in war is prohibited.

Example: During World War I, the use of poison gas by both the German and British armies was prohibited under the Second Hague Convention. However, the German army still used poison gas and was criticized for its use.

3. The Fourth Hague Convention: The Fourth Hague Convention was signed in 1907 and established the principle that the use of prohibited weapons in war is prohibited. It was signed by 33 countries and established the principle that the use of prohibited weapons in war is prohibited.

Example: During World War II, the use of chemical weapons by both the German and Soviet armies was prohibited under the Fourth Hague Convention. However, the Soviet army used chemical weapons against the Polish army in 1939, which was not a violation of the Hague Convention.

$\cdots$

These are just a few examples of how the Hague Conventions have been used to protect the rights of civilians in international armed conflicts. The Hague Conventions have been ratified by more than 190 countries and have been instrumental in shaping international law and protecting the rights of civilians in armed conflicts. & 67 & 63.4 \\
\midrule
\factalign & The Hague Conventions are a series of international treaties that were signed in the Netherlands city of The Hague in 1899. These conventions were designed to address the issue of international law and to promote cooperation between different countries.

The Hague Conventions are divided into several different treaties, each addressing a different aspect of international law. Some of the most significant Hague Conventions include:

1. The Hague Convention for the Protection of Cultural Property in the Event of Armed Conflict (1954)
2. The Hague Convention Respecting the Laws and Customs of War on Land (1907)
3. The Hague Convention for the Protection of Victims of International Armed Conflicts (1977)
4. The Hague Convention on the Civil Aspects of International Litigation (1971)
5. The Hague Convention on the International Sale of Goods (1980)
6. The Hague Convention on the Protection of the Rights of Visigothic Migrants (1998)

These conventions were signed by a large number of countries, and they have had a significant impact on international law. For example, the Hague Convention for the Protection of Cultural Property in the Event of Armed Conflict was signed in 1954, and it has been ratified by a large number of countries. This convention has been used in a number of different situations, including the conflict in Bosnia and Herzegovina.

$\cdots$

In terms of specific examples, the Hague Conventions have been used in a number of different situations. For example, the Hague Convention for the Protection of Cultural Property in the Event of Armed Conflict has been used in a number of different conflicts, including the conflict in Bosnia and Herzegovina. The Hague Convention Respecting the Laws and Customs of War on Land has been used in a number of different situations, including the conflict in Afghanistan.

Overall, the Hague Conventions are considered to be some of the most significant international treaties of the 20th century. They have had a significant impact on international law, and they continue to be relevant today. & 65 & 78.9 \\
\bottomrule
\end{tabular}
\caption{An example of model generations. The generations are cut short due to space limit.}
\label{tab:example}
\end{table*}

\section{Qualitative Examples}
\label{app:example}
We include a qualitative in Table~\ref{tab:example}.

\end{document}